\documentclass[10pt,twocolumn,letterpaper]{article}

\usepackage[final]{cvpr}

\usepackage{times}
\usepackage{latexsym}
\usepackage[T1]{fontenc}
\usepackage[utf8]{inputenc}
\usepackage{microtype}
\usepackage{graphicx}
\usepackage{amsmath}
\usepackage{booktabs}
\usepackage{multirow}
\usepackage{framed}
\usepackage{xcolor}
\usepackage{enumitem}
\setlist[itemize]{leftmargin=*,topsep=2pt,itemsep=1pt}

\usepackage{tikz}
\usetikzlibrary{shapes.geometric,arrows.meta,positioning,backgrounds,calc}

\usepackage[pagebackref,breaklinks,colorlinks,allcolors=blue]{hyperref}

\def\paperID{22}
\def\confName{CVPR}
\def\confYear{2026}

\title{Zero-Shot Traffic Accident Detection via a Coarse-to-Fine VLM-Tracking Pipeline}

\renewcommand*{\thefootnote}{\fnsymbol{footnote}}

\author{%
Dipit Saha\thanks{Equal contribution.}, Shah Mohammad Abdul Mannan\footnotemark[1], Mohammad Raihan Rashid\footnotemark[1],\\
Ruwad Naswan\footnotemark[1], Ahnaf Tahmid\footnotemark[1]\\
Bangladesh University of Engineering and Technology\\
{\small Correspondence: \tt sdipit099@gmail.com}%
}

\begin{document}
\maketitle
\renewcommand*{\thefootnote}{\arabic{footnote}}
\setcounter{footnote}{0}

\begin{abstract}
Traffic surveillance cameras capture accidents continuously, yet converting raw CCTV footage into structured event records that pinpoint \textit{when}, \textit{where}, and \textit{what type} of collision occurred remains unsolved at scale. The ACCIDENT @ CVPR benchmark evaluates exactly this joint prediction under a strict constraint: no labeled real-world training data is available. We introduce a training-free, two-pass coarse-to-fine pipeline that pairs a frozen Qwen3-VL-32B-Instruct vision-language model with YOLO11x object detection and BoT-SORT tracking. A first pass sparsely samples the full clip to anchor the collision moment in time; a second pass re-examines a tight window around that estimate using frames annotated with stable vehicle identities and normalized bounding-box coordinates, which gives the model both a visual overlay and an explicit numeric description of the same scene. On the official 2{,}027-clip real-CCTV test set, our system achieves a three-way harmonic mean score of 0.504, surpassing all organizer-published baselines including the best multi-model ensemble (0.412) by a 22\% relative margin.
\end{abstract}

\section{Introduction}
\label{sec:intro}

Traffic surveillance cameras capture accidents continuously, yet extracting
structured event records from raw footage, predicting \emph{when}, \emph{where},
and \emph{what type} of collision occurred, remains an open problem at scale.
CCTV footage is unforgiving: low resolution, heavily compressed, often filmed at
night, with vehicles that may span only a handful of pixels.
The ACCIDENT @ CVPR benchmark~\cite{picek2026accident,accidentbench2026challenge}
targets precisely these conditions, framing accident understanding as a joint
three-way task and evaluating submissions through the harmonic mean of all three
branches, a metric that collapses to zero if any single branch fails.

The central difficulty is the absence of real labeled training data.
Competitors have access only to a CARLA-based synthetic development set;
all systems must generalize zero-shot to real CCTV clips at test time.
A vision-language model~(VLM) seems a natural fit for this setting, given
its broad visual and semantic priors. Yet a naive single-pass call over the
full clip fails on all three axes. Accidents occupy roughly 100--500\,ms of
a 30\,s clip, so uniform 4\,fps sampling routinely misses the contact moment.
Low-resolution pixels make precise coordinate recovery unreliable even when
the right frame is found. And under uncertainty, the model exhibits a
consistent \emph{mid-video temporal bias}, defaulting to the clip midpoint
$\hat{t} \approx D/2$ for a clip of duration $D$, regardless of where the
event actually occurs.
These are not incidental failure modes; each one independently collapses
the harmonic mean score.

We address them through a training-free two-pass pipeline built around a
frozen Qwen3-VL-32B-Instruct model. A \emph{coarse pass} sparsely samples
the full clip to anchor the collision in time, sidestepping the contact-frame
problem by narrowing the search window before any spatial reasoning begins.
A \emph{fine pass} then re-examines a tight window around that estimate,
running YOLO11x and BoT-SORT at native frame rate to maintain stable vehicle
identities across frames. Each selected frame is annotated with bounding boxes
and vehicle labels, while the same detections are simultaneously written as
normalized coordinate text, giving the model visual and numeric views of the
scene that are harder to recover from low-resolution pixels alone.
Three inference-time safeguards complete the design: a sub-second temporal
floor, an out-of-memory fallback chain, and a type-continuity hint carried
between passes; Table~\ref{tab:robustness} summarizes the failure modes they are
designed to prevent. Our system surpasses all organizer-published baselines, including
the best multi-model ensemble, by a 22\% relative margin in harmonic mean
score, without any real-data fine-tuning.

\section{Related Work}
\label{sec:related}

\paragraph{Pretrained multimodal models for video understanding.}
Recent vision-language models have improved multi-frame reasoning and
open-ended video understanding. Qwen2.5-VL introduces dynamic frame-rate
sampling and stronger temporal grounding~\cite{qwen25vl,su2024rope},
while Qwen3-VL further improves long-context visual reasoning~\cite{qwen3vl2025}.
Other pretrained video models such as Video-LLaMA~\cite{li2023videollama}
and V-JEPA~2~\cite{vjepa2} also show that large-scale pretraining can
support temporal understanding without task-specific supervision.

\paragraph{Accident and anomaly understanding in surveillance video.}
Earlier surveillance-based accident detection methods rely on motion cues,
optical flow, or hand-crafted collision heuristics~\cite{picek2026accident,teed2020raft}.
More recent work uses large multimodal models for training-free or weakly
supervised anomaly reasoning. Holmes-VAD~\cite{holmesvad2024} and
LAVAD~\cite{lavad2024} show that captioning and language-based reasoning
can be effective without end-to-end task-specific training. Other methods,
including CrashSight~\cite{crashsight2026} and STER-VLM~\cite{stervlm2025},
move closer to accident understanding in traffic scenes, but they rely on
fine-tuning or more specialized adaptation. Our setting is stricter: we
target zero-shot accident understanding on real CCTV video without real-data
fine-tuning.

\paragraph{Detection, tracking, and visual guidance.}
Object detection and multi-object tracking remain useful for structuring
crowded traffic scenes before higher-level reasoning. We build on YOLO11
for vehicle detection~\cite{yolo11} and BoT-SORT for identity-consistent
tracking across frames~\cite{aharon2022botsort}. We also draw on prior work
showing that visual cues added directly to images can influence VLM
predictions~\cite{shtedritski2023whatredcircle}. Our use of annotated frames
follows this idea, but pairs visual overlays with normalized coordinate text
so that the model receives both image-level and explicit geometric evidence.




\section{Task and Metric}
\label{sec:task}

Given a fixed-view CCTV clip, the task is to predict the accident
time $\hat{t}$, the normalized impact location $(\hat{x}, \hat{y})$,
and the collision type
$\hat{c} \in \{$\texttt{head-on, rear-end, sideswipe, t-bone, single}$\}$.

Evaluation is performed on
three components: temporal localization, spatial localization, and
collision-type classification. The temporal score $\mathcal{T}$ and
spatial score $\mathcal{S}$ use Gaussian-style similarity measures,
while the classification score $\mathcal{C}$ is exact-match accuracy.
The official unified benchmark score is the harmonic mean of the three:
\begin{equation}
\mathrm{ACCIDENT\ Score} =
\frac{3}{\mathcal{T}^{-1} + \mathcal{S}^{-1} + \mathcal{C}^{-1}}.
\label{eq:accident_score}
\end{equation}
This scoring strongly penalizes weak performance on any single branch.

In the zero-shot track, all 2,027 real-CCTV clips are held for
evaluation only. The sole labeled resource available
is a 2,211-clip synthetic set rendered in
CARLA~\cite{picek2026accident,accidentbench2026challenge}.

\section{System Overview}
\label{sec:system}

\begin{figure*}[t]
\centering
\includegraphics[width=\textwidth]{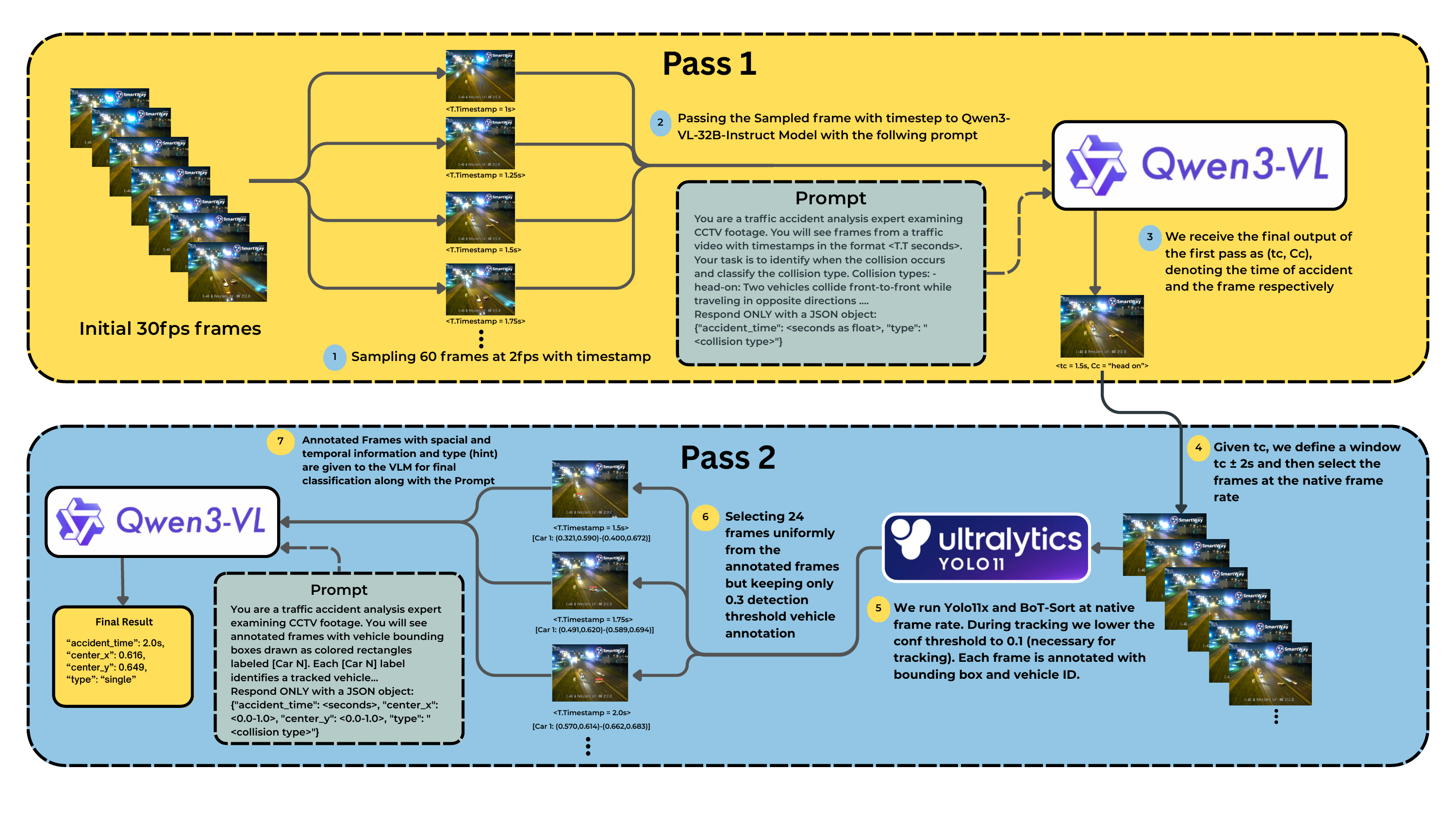}
\caption{Compact two-pass pipeline. Pass~1 performs temporal localization on sparse timestamped frames to produce a coarse time/type anchor; Pass~2 restricts attention to a $\pm$2\,s window, adds YOLO11x + BoT-SORT tracking and coordinate text, and returns the final structured prediction.}
\label{fig:pipeline}
\end{figure*}

Figure~\ref{fig:pipeline} shows the full pipeline. For each clip, the system
runs two sequential passes of the video VLM Qwen3-VL-32B-Instruct.

\subsection{Pass 1: Coarse Temporal Localization}
\label{sec:coarse}

We sample up to 60 frames at 2 fps from the full
clip. Frames are then passed to the VLM interleaved with timestamp
tokens in the official Qwen format, i.e., the token \texttt{<T.T seconds>} after each image.
The model is then instructed to return only a JSON
object \texttt{\{"accident\_time": <time>, "type": <type>\}},
denoted by $(\hat{t}_c, \hat{c}_c)$.
We use the coarse time prediction $\hat{t}_c$ to center Pass~2 and the coarse
collision-type prediction $\hat{c}_c$ as a soft type hint.

\subsection{Pass 2: Fine Spatial and Temporal Refinement}
\label{sec:fine}

Given $\hat t_c$, we define a 4-second refinement window
$[\hat t_c - 2\,\mathrm{s},\, \hat t_c + 2\,\mathrm{s}]$ and run
YOLO11x~\cite{enosyolo11x} in Ultralytics tracking mode with the default
BoT-SORT configuration~\cite{aharon2022botsort} at the video's native frame
rate. A low tracking threshold (\texttt{conf}=0.1) keeps vehicles that turn
blurry, small, or partly occluded. Running at native fps matters: tracking on
sub-sampled frames swaps IDs between frames, which weakens both the visual
prompt and the final prediction.

We then uniformly sample 24 frames from the tracked window and annotate them
with bounding boxes and vehicle IDs using the \texttt{BoxAnnotator} and
\texttt{LabelAnnotator} utilities from the \texttt{supervision}
library~\cite{supervision2024}. A separate visualization threshold of $0.3$
on YOLO confidence keeps the main collision participants while suppressing
low-confidence background boxes.

Each annotated frame is paired with a timestamp token \texttt{<T.T seconds>}
so the model can place it inside the refinement window, and the detected
vehicles are also listed as normalized bounding boxes:
\begin{center}
\small
\verb|[Car 1: (0.491,0.620)-(0.589,0.694)]|
\end{center}
with $(x_1,y_1)$ and $(x_2,y_2)$ the top-left and bottom-right corners in
$[0,1]$. The coarse-pass type prediction $\hat{c}_c$ is passed in as a soft
hint, and the model is asked to return a single JSON object
\texttt{\{"accident\_time", "center\_x", "center\_y", "type"\}}.

\begin{table}[ht]
\centering
\small
\setlength{\tabcolsep}{4pt}
\begin{tabular}{p{1.65cm}p{2.05cm}p{3.0cm}}
\toprule
\textbf{Mechanism} & \textbf{Trigger} & \textbf{Failure prevented} \\
\midrule
Temporal floor & $\hat{t}<0.3$\,s from either pass & removes degenerate zero-time outputs that would collapse $\mathcal{T}$ \\
YOLO retry & detector OOM at \texttt{imgsz}=1088 & preserves fine-pass spatial reasoning on long clips \\
Raw-frame fallback & zero detections in fine window & avoids full pass failure on night or compressed clips \\
Short-window retry & VLM OOM in Pass~2 & recovers inference with fewer frames and tighter context \\
JSON safeguard & malformed structured output & preserves non-zero prediction rather than crashing the clip \\
\bottomrule
\end{tabular}
\caption{Inference-time safeguards used to prevent catastrophic
harmonic-mean failures.}
\label{tab:robustness}
\end{table}

\section{Experiments}
\label{sec:experiments}

\paragraph{Setup.}
All inference runs on a single NVIDIA H100 80\,GB GPU (bfloat16, FlashAttention-2~\cite{dao2022flashattention})
using YOLO11x and frozen Qwen3-VL-32B-Instruct weights; no model is trained or
fine-tuned on ACCIDENT labels.

\paragraph{Main results.}
Table~\ref{tab:results} compares our system against organizer-published
baselines on the official real-CCTV test set. The first four rows are
organizer-reported numbers, and the last row is our system, scored on the full
2{,}027-clip test set by the organizer's category-wise evaluator hosted on a
separate official page; the Kaggle-style leaderboard itself reports only a
combined HM (public $0.499$, private $0.503$).

\begin{table}[ht]
\centering
\small
\setlength{\tabcolsep}{3pt}
\begin{tabular}{lcccc}
\toprule
\textbf{System} & $\mathcal{T}$ & $\mathcal{S}$ & $\mathcal{C}$ & \textbf{HM} \\
\midrule
Heuristics (no classification)~\cite{picek2026accident}
  & 0.287 & 0.273 & 0.000 & 0.000 \\
Naive Baseline~\cite{picek2026accident}
  & 0.190 & 0.250 & 0.335 & 0.245 \\
Molmo-7B~\cite{picek2026accident}
  & 0.343 & 0.488 & 0.293 & 0.358 \\
Best from all~\cite{picek2026accident}
  & 0.343 & 0.488 & 0.433 & 0.412 \\
\midrule
\textbf{Our System} &
  \textbf{0.549} & \textbf{0.468} & \textbf{0.503} & \textbf{0.504} \\
\bottomrule
\end{tabular}
\caption{Official test-set results. The first four rows are organizer-published
baselines~\protect\cite{accidentbench2026challenge}, and \textbf{HM} is the
three-way harmonic mean. Our row is the organizer's category-wise evaluation
over all 2{,}027 clips; the Kaggle leaderboard records the combined HM as
$0.499$ (public) and $0.503$ (private).}
\label{tab:results}
\end{table}

\paragraph{Interpreting the branch gains.}
Relative to \emph{Best from all}, our system gains $+0.206$ on $\mathcal{T}$ and
$+0.070$ on $\mathcal{C}$, while $\mathcal{S}$ regresses by $0.020$. Inspecting
the clips behind that regression shows two recurring cases. A vehicle entering
the frame late, such as the cross-traffic car in a t-bone, is dropped by
BoT-SORT before it clears the $0.3$ visualization threshold; the fine pass then
anchors on the larger, stably tracked vehicle, which sits one to two
car-lengths from the true contact point. Night and heavy compression also
suppress YOLO confidence across the scene, the fine window falls back to raw
frames, and the model regresses toward $(0.5,0.5)$ instead of committing. Both
trade a small spatial error for the larger temporal and classification gains
that tracking and coordinate text provide.

\paragraph{Engineering progression on official test set.}
Because real-CCTV labels are hidden, Table~\ref{tab:ablation} traces
submission-level changes as the closest substitute for a controlled ablation.
The dominant gain ($+0.234$ HM) came from replacing the organizer baseline
with a single-pass VLM call; subsequent design choices each contributed
smaller but consistent improvements.

\begin{table}[ht]
\centering
\small
\setlength{\tabcolsep}{3pt}
\begin{tabular}{p{2.6cm}p{2.95cm}cc}
\toprule
\textbf{Variant} & \textbf{Progressive change} & \textbf{Pub.} & \textbf{Priv.} \\
\midrule
Organizer Naive Baseline~\cite{picek2026accident}
  & organizer-reported reference & 0.226 & 0.246 \\
Single-pass VLM
  & full-clip VLM in place of heuristics & 0.469 & 0.479 \\
+ Coarse temporal loc.
  & coarse accident-time anchoring & 0.481 & 0.491 \\
+ Spatial grounding
  & YOLO boxes in the fine stage & 0.472 & 0.492 \\
+ Temporal refinement
  & tighter window around first contact & 0.493 & 0.499 \\
+ Two-pass coarse-to-fine (\textbf{ours})
  & balanced coarse-to-fine config & \textbf{0.499} & \textbf{0.504} \\
\bottomrule
\end{tabular}
\caption{Engineering progression on the hidden test server. Each row adds one
design change; \textbf{Pub.}/\textbf{Priv.} are the public- and
private-leaderboard HMs for that submission. The final row matches the
full-evaluator HM of $0.504$ reported in Table~\ref{tab:results}; the
corresponding raw private-leaderboard value is $0.50265$. Because these come
from successive hidden-test submissions, the table is supportive development
evidence rather than a controlled ablation.}
\label{tab:ablation}
\end{table}

\paragraph{Reproducibility and inference budget.}
Each clip uses two VLM calls: 60 frames at 2\,fps (coarse pass) and
24 annotated frames over a $\pm2$\,s window (fine pass).
End-to-end latency is 40--90\,s/clip on an H100.

\paragraph{Error analysis.}
Manual inspection of clips that triggered fallback behavior or produced
suspicious coordinates revealed four dominant failure patterns,
summarized in Table~\ref{tab:failures}.

\begin{table}[ht]
\centering
\small
\setlength{\tabcolsep}{4pt}
\begin{tabular}{p{1.8cm}p{0.9cm}p{4.35cm}}
\toprule
\textbf{Failure mode} & \textbf{Main hit} & \textbf{Observed behavior} \\
\midrule
Night + compression & $\mathcal{S}$ &
YOLO often misses vehicles; the fine pass falls back to raw frames and
frequently regresses toward $(0.5,0.5)$. \\
Sideswipe & $\mathcal{T}, \mathcal{C}$ &
First contact is subtle, so the coarse pass can lock onto a near-miss
frame instead of the impact moment. \\
Single-vehicle & $\mathcal{C}$ &
The VLM sometimes hallucinates a second vehicle and misclassifies the
event as \texttt{t-bone}. \\
Mid-video bias & $\mathcal{T}$ &
Under high uncertainty, the VLM defaults toward $\hat{t} \approx D/2$. \\
\bottomrule
\end{tabular}
\caption{Dominant failure patterns identified from manual inspection of
clips that triggered fallback behavior or produced suspicious coordinates.}
\label{tab:failures}
\end{table}

\paragraph{Practical lessons.}
(i)~\texttt{do\_sample=True} at V1 caused JSON drift and center-clustered
22\% of clips; switching to greedy decoding eliminated this.
(ii)~Including all YOLO detections diluted collision evidence; filtering
to track-stable, highest-confidence boxes reduced out-of-range outputs.
(iii)~Equal temporal density in both passes wastes the fine pass:
narrowing from $\pm4$\,s at 16\,fps to $\pm2$\,s at 24\,fps yielded the
largest $\mathcal{T}$ gain.
(iv)~BoT-SORT on subsampled frames caused ID swaps
(\texttt{[Car 4]}$\to$\texttt{[Car 9]}); native-fps tracking preserved
vehicle identity across the fine window.

\paragraph{Limitations.}
The pipeline requires two large-VLM calls plus detector-tracker inference per
clip, and the absence of labeled real-CCTV data leaves some design choices
empirically underspecified.

\section{Conclusion}
\label{sec:conclusion}

We presented a training-free two-pass pipeline for joint temporal,
spatial, and collision-type prediction in CCTV traffic accidents.
Our method combines a frozen Qwen3-VL-32B-Instruct model with
YOLO11x detection and BoT-SORT tracking, and uses both annotated
frames and normalized coordinate text to improve fine-grained reasoning.
On the test dataset, it achieves a harmonic mean
score of 0.504, exceeding the best organizer-published baseline by \textbf{22\%}
relative.

Future work should explore: a lightweight temporal pre-filter (e.g.,
V-JEPA~2 surprise signal or RAFT-jerk curve) to narrow the coarse
window; a dedicated single-vehicle detector to reduce \texttt{t-bone}
misclassifications; and targeted domain adaptation using the synthetic
CARLA labels, which we deliberately avoided to preserve the zero-shot
design.

{\small
\bibliographystyle{ieeenat_fullname}
\bibliography{custom}
}

\clearpage
\onecolumn

\end{document}